# Study of the Dynamic Coupling Term (μ) in Parallel Force/Velocity Actuated Systems

Dinesh Rabindran and Delbert Tesar
Robotics Research Group, Department of Mechanical Engineering
University of Texas at Austin

*Abstract*—Presented in this paper is an actuator concept, called a Parallel Force/Velocity Actuator (PFVA), that combines two fundamentally distinct actuators (one using low gear reduction or even direct drive, which we will call a Force Actuator (FA) and the other with a high reduction gear train that we will refer to as a Velocity Actuator (VA)). The objective of this work is to evaluate the effect of the relative scale factor, RSF, (ratio of gear reductions) between these inputs on their dynamic coupling. We conceptually describe a Parallel Force/Velocity Actuator (PFVA) based on a Dual-Input-Single-Output (DISO) epicyclic gear train. We then present an analytical formulation for the variation of the dynamic coupling term w.r.t. RSF. Conclusions from this formulation are illustrated through a numerical example involving a 1-DOF four-bar linkage. It is shown, both analytically and numerically, that as we increase the RSF, the two inputs to the PFVA are decoupled w.r.t. the inertia torques. This understanding can serve as an important design guideline for PFVAs. The paper also presents two limitations of this study and suggests future work based on these caveats.

## I. INTRODUCTION

Dexterous manipulation applications require very smooth motion planning and intricate force-profile management. The most challenging tasks are the ones in which force and motion have to be managed in the same direction, like deburring. Dexterous tasks can, in the limit, be classified into two mutually exclusive functional regimes, namely, force-controlled and velocity controlled. In purely force-controlled tasks, the objective is to achieve a desired interaction force (velocity management being secondary) and in purely velocity-controlled tasks, the goal is to adhere to a reference motion plan (force control being secondary).

The Electro-Mechanical Actuators (EMA) that drive intelligent mechanical systems (like robots) can also be classified into "ideal" Force Actuators (FA) and "ideal" Velocity Actuators (VA). The reduction used in their gear trains characterizes them as FA or VA. A high reduction gear ratio (such as 150:1) makes the actuator behave like a velocity generator or VA in that it can manage a commanded velocity while resisting force disturbances with minimal impact on the commanded velocity. On the other hand, an EMA with a low reduction gear ratio (such as a direct drive actuator) acts like an ideal force generator or a FA. In other words, an FA can maintain a reference force while reacting to velocity disturbances. These inverse characteristics arise due to the fact that force and velocity are power conjugate variables.

In this work, we propose an approach that combines an FA and a VA within the same design in parallel. Further we focus our attention here on a study of the effect of relative scale factor between the two inputs, on the dynamic coupling between them. This study is important because the dynamic (or inertial) coupling between the inputs dictates how much each of them gets disturbed by the variations in the inertia at the system output. A clear analytical and physical understanding of this coupling term will thus be a significant step towards the design of such systems.

The paper is organized to present a review of the pertinent literature first. We then present the concept of the Parallel Force/Velocity Actuator (PFVA) and its analysis. Subsequently, we present an analytical study of the dynamic coupling term as a function of the relative scale factor between the two inputs to the PFVA. A numerical example is included to illustrate the conclusions from the analytical formulation. The paper concludes with a discussion of the results and two suggestions for future work.

## II. LITERATURE REVIEW

The fundamental issue in this paper is to characterize the dynamic coupling between the two inputs in a Parallel Force/Velocity Actuator (PFVA) [15] which will then serve as a design guideline. The objective of the PFVA is to obtain a variety of dynamic responses at the system's output, from the highly stiff 'pure velocity controlled response' to the more forgiving 'pure force controlled response'. In this section we will review relevant actuation concepts that were motivated by similar goals.

The significance of the actuator and its properties in determining the limits of performance of a robotic system have been recognized earlier by Tesar [1] and Hollerbach *et al*. [2]. The dependence of system performance on actuator characteristics (especially bandwidth) has been studied in the past by Eppinger and Seering [3] and Hollerbach *et al*. [2]. To improve force-controlled performance, research at the MIT Leg Lab purposely included series compliance between the actuator and the load (Series Elastic Actuation, SEA) to reduce impact loads [4]. Joint level torque-control was proposed by Vischer and Khatib [5] to reduce the non-linear effects in actuators. This method can actively change the response of an actuator (very stiff or forgiving) depending on the task; but for frequencies beyond the joint torque control bandwidth, the response is governed by its structural compliance. Joint torque control is also used in the operation of the DLR-III lightweight manipulator arm [6]. A new actuation concept, called Distributed Macro-Mini (DM$^2$) Actuator, for human-centered robotic systems, was proposed at the Stanford Robotics Lab [7]. This research was driven by the need to design safer as well as better performing robots. The central idea of DM$^2$ was to partition input torque generation into high frequency and low frequency components that sum in parallel and are appropriately located at the joint and the base of the manipulator. At the University of Texas, the actuation effort has been toward maximizing the number of choices (in the force and motion domains) available within the actuator. This includes dual-level control for fault-tolerance [8] and layered control [9]. The Parallel Coupled Micro-Macro Actuator (PaCMMA) from MIT [10] was a concept similar to layered control [9]. The objective of PaCMMA, an in-parallel design, was to improve the force resolution and closed-loop force control bandwidth; however it has packaging issues due to its complexity. An actuation mechanism was proposed by Kim *et al.* [11] at Korea University based on a planetary gear train. This is a Dual Actuator Unit (DAU) that is driven by two sub-systems, a "positioning actuator" and a "stiffness modulator". The DAU concept bears resemblance to the Force/Motion Actuator[1] proposed earlier by Tesar [12]. The DAU operates such that the "stiffness modulator" biases the position of the "positioning actuator" when a collision is detected. It is well-packaged in a protoype; however, Kim *et al.* [11] have not investigated any of the operational issues associated with multi-input gear trains (for example, the dynamic influence of one input on the other). Another actuation paradigm based on a Continuously Variable Transmission (CVT) was proposed by Faulring *et al.* [13] at Northwestern University which they called *Cobotics*. The goal of this CVT approach was to improve power efficiency.

The premise of the PFVA concept is that we could dynamically "mix" the contributions of a pair of low-reduction (Force Actuator, FA) and high-reduction (Velocity Actuator, VA) actuators in-parallel to obtain a variety of responses at the system's output. This concept of Parallel Force/Velocity actuation revolves around the fact that the gear-ratio is a significant property of the actuator [14]. In this paper, however, we restrict our study to the issue of dynamic coupling between the inputs to the PFVA. In the next section we will present the concept of a Parallel Force/Velocity Actuator (PFVA) and in subsequent sections we will lay out the analytics required to characterize the dynamic coupling term between the constituent sub-systems, namely, the FA and VA.

---

[1] Conceptual origin of the PFVA (proposed in this paper)

## III. PARALLEL FORCE/VELOCITY ACTUATION PRINCIPLE

In this section we will present the Parallel Force/Velocity Actuation (PFVA) concept. Recognizing that the gear ratio of an actuator is a significant property that influences the dynamic response at the output of the system, we will classify EMAs into two mutually exclusive classes based on their transmission ratios (Fig 1), namely the ideal "Force Actuator" (FA) and the ideal "Velocity Actuator" (VA). The FA is a low-reduction actuator (10-15 to 1 or even direct drive) that is ideal for tasks in which the output force is being controlled while velocity disturbances are being tolerated. In other words, a FA is a perfect force/torque source. On the other hand, the VA is a high-reduction actuator (100-150 to 1) that performs well in applications where the velocity needs to be controlled precisely while rejecting force disturbances. A VA is an ideal velocity source. The PFVA concept combines the above two classes of actuators in one Dual Input Single Output (DISO) design using an epicyclic gear train.

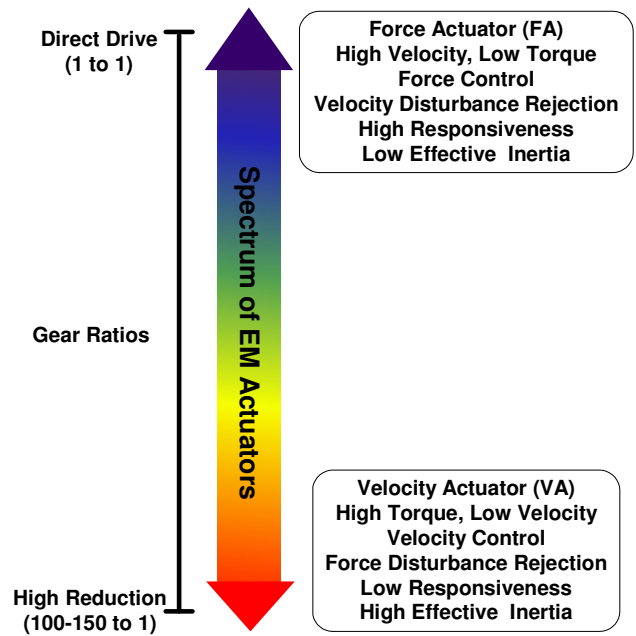

Fig 1. Spectrum of EMAs based on Gear Reduction

### A. Parallel Force/Velocity Actuation – Analysis

In a Parallel Force/Velocity Actuator (PFVA), we combine a Force Actuator and a Velocity Actuator in parallel using a 2-DOF epicyclic gear train. In Fig 2 is shown the configuration of the epicyclic gear train to realize our objective of incorporating fundamentally distinct force and motion priorities within the same actuator. There are two inputs to the epicyclic gear train. As shown in the schematic (Fig 2), the prime-mover of the FA drives the sun-gear shaft and that of the VA drives the carrier. The output is the ring gear. In keeping with our nomenclature for the ideal actuators, the prime-movers for the FA and VA will respectively be called the *force input* (or force sub-system, FSS) and *velocity input* (or velocity sub-system, VSS). The properties of a Parallel Force/Velocity Actuator are listed in Table I.

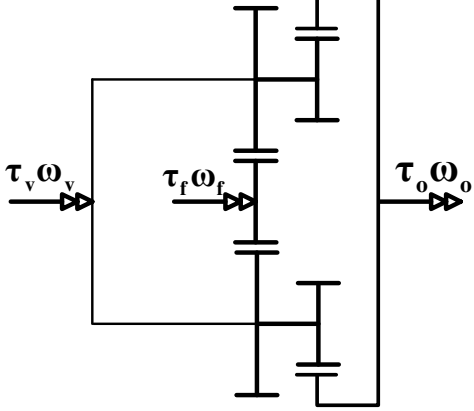

Fig 2. Schematic of a Dual Input Single Output Planetary Gear Train.

TABLE I
PARALLEL FORCE/VELOCITY ACTUATOR PROPERTIES

| Symbol | Quantity |
|---|---|
| $\omega_o$ | Output Angular Velocity of the PFVA |
| $\omega_v$ | Angular Velocity of the Velocity Input in PFVA |
| $\omega_f$ | Angular Velocity of the Force Input in PFVA |
| $\tau_o$ | Output Torque of the PFVA |
| $\tau_v$ | Torque at the Velocity Input in PFVA |
| $\tau_f$ | Torque at the Force Input in PFVA |
| $R_f$ | Gear Reduction for Force Side |
| $R_v$ | Gear Reduction for Velocity Side |
| $\rho$ | Relative scale factor Between Sub-Systems, $\dfrac{R_f}{R_v}$ |

In the following analysis we will use some results from Muller [16]. The output velocity of the PFVA is a linear combination of the input velocities with scaling dependent on the gear ratios of their respective force paths. We may express the velocity mapping as a velocity summation:

$$\omega_o = R_v \omega_v + R_f \omega_f \quad (1)$$

Now, considering the conservation of power (if we assume no power loss due to inefficiency),

$$\tau_o \omega_o = \tau_v \omega_v + \tau_f \omega_f \quad (2)$$

Using Eqns. (1) and (2) it can be shown that the torque mapping between the inputs and the output of the PFVA is as follows:

$$\tau_o = \frac{\tau_v}{R_v} = \frac{\tau_f}{R_f} \quad (3)$$

The above relation may be re-written in vector form as follows:

$$\begin{bmatrix} \tau_v \\ \tau_f \end{bmatrix} = \begin{bmatrix} R_v \\ R_f \end{bmatrix} \tau_o \quad (4)$$

It is a property of the epicyclic gear train [16] shown in Fig 2, that:

$$R_v + R_f = 1 \quad (5)$$

Let us define the Relative scale factor (ρ) between the inputs as follows:

$$\rho = \frac{R_f}{R_v} \quad (6)$$

Then we may use the relation in Eqn. (5) to re-write the individual gear reductions as follows:

$$R_v = \frac{1}{\rho + 1} \quad (7)$$

$$R_f = \frac{\rho}{\rho + 1} \quad (8)$$

From Eqns.(7) and (8), we recognize that $\rho = 1$ is a singularity for the epicyclic gear train. The physical meaning of this scenario is that the whole system rotates with one single velocity and behaves like a Single Input Single Output (SISO) transmission system with the gear ratio equal to unity.

IV. STUDY OF DYNAMIC COUPLING TERM ( $\mu$ )

In this section we will present our study of the dynamic coupling term (μ) in the input reflected inertia matrix of a 1-DOF PFVA actuated system. This term (μ) is an important design metric for the DISO Parallel Force/Velocity Actuator (PFVA). Browning and Tesar [18] recognized the importance of this coupling term as a performance criterion for operating n-DOF manipulator systems. In the following derivation, however, we will consider a single-link equivalent of a non-linear single-DOF mechanism (Fig 3). The joint displacement variable is θ, and those for the two inputs are φ_V and φ_F (respectively for the velocity subsystem and the force-subsystem). Transformation of the nonlinear dynamics of a complex system to the single-link equivalent is discussed in [19]. As will become evident from the following derivation, the fundamental characteristic of μ is independent of the reflected inertia at the joint (output of the PFVA). It is a function only of the Relative Scale Factor (RSF) (or ρ).

The velocity transformation from the PFVA input space to the joint space may be written in matrix form as follows.

$$\dot{\theta} = \begin{bmatrix} \mathbf{G}_\phi^\theta \end{bmatrix} \dot{\boldsymbol{\phi}} = \begin{bmatrix} R_v & R_f \end{bmatrix} \begin{bmatrix} \dot{\phi}_v \\ \dot{\phi}_f \end{bmatrix} \quad (9)$$

Here, $\left[\mathbf{G}_\phi^\theta\right]$ represents a (constant) matrix of kinematic influence coefficients (Refer [17]) consisting of the velocity ratios for the two inputs to the output. If $\mathbf{I}_{\theta\theta}^*$ is the joint reflected inertia of a 1-DOF mechanism that is driven by the PFVA, then we can reflect this inertia to the two inputs ($\mathbf{I}_{\phi\phi}^*$) of the PFVA, namely the VA and FA as follows:

$$\mathbf{I}_{\phi\phi}^* = \mathbf{I}_M + \left[\mathbf{G}_\phi^\theta\right]^T I_{\theta\theta}^* \left[\mathbf{G}_\phi^\theta\right] \quad (10)$$

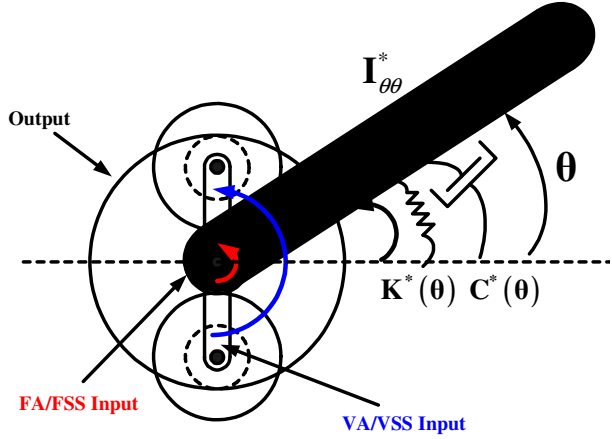

Fig 3. Single link equivalent of a complex 1-DOF nonlinear mechanism. Refer [19] for details.

Here, $\mathbf{I}_M = \begin{bmatrix} I_{Mv} & 0 \\ 0 & I_{Mf} \end{bmatrix}$ is a diagonal matrix of the rotor inertias of the motors driving the two inputs. Recognizing from Eqns. (7) and (8) that,

$$\left[\mathbf{G}_\phi^\theta\right] = \begin{bmatrix} \dfrac{1}{1+\rho} & \dfrac{\rho}{1+\rho} \end{bmatrix} \quad (11)$$

it can be shown that;

$$\mathbf{I}_{\phi\phi}^* = \begin{bmatrix} I_{Mv} + I_{\theta\theta}^* \dfrac{1}{(\rho+1)^2} & I_{\theta\theta}^* \dfrac{\rho}{(\rho+1)^2} \\ I_{\theta\theta}^* \dfrac{\rho}{(\rho+1)^2} & I_{Mf} + I_{\theta\theta}^* \dfrac{\rho^2}{(\rho+1)^2} \end{bmatrix} \quad (12)$$

We now have an explicit formula for the dynamic coupling term as a function of the Relative Scale Factor[2] [$\mu(\rho)$]:

$$\mu = I_{\theta\theta}^* \dfrac{\rho}{(\rho+1)^2} \quad (13)$$

To understand the sensitivity of the dynamic coupling ($\mu$) to design changes in the RSF ($\rho$), we take differentials on both sides of Eqn. (13).

[2] RSF – Relative Scale Factor ($\rho$)

$$\Delta\mu = \left[I_{\theta\theta}^* \dfrac{1-\rho}{(\rho+1)^3}\right]\Delta\rho \quad (14)$$

The relations developed in Eqns. (13) and (14) have been plotted in Fig 4. A limit analysis can be done on Eqns. (13) and (14) to recognize that as $\rho\to\infty$, $\mu\to 0$ and $(\Delta\mu/\Delta\rho)\to 0$. This is important design knowledge and can be used as a design rule of thumb for PFVA actuated mechanisms.

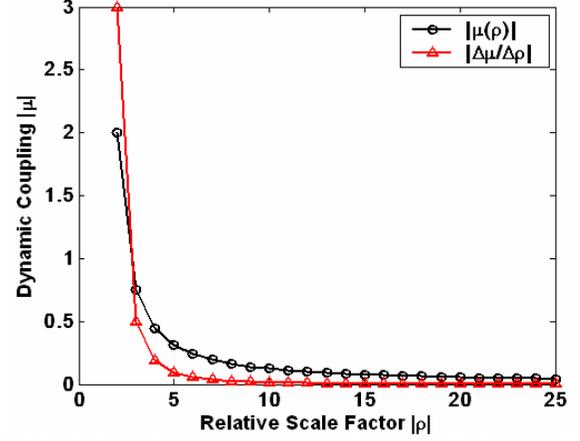

Fig 4. Variation of dynamic coupling term ($\mu$) and its derivative w.r.t. the Relative Scale Factor (RSF or $\rho$). As $\rho\to\infty$, $\mu\to 0$ and $(\Delta\mu/\Delta\rho)\to 0$. $I_{\theta\theta}^*$ is assumed to be unity in this plot.

A limit analysis similar to the one suggested above can be performed on the reflected inertia term, $\mathbf{I}_{\phi\phi}^*$, to result in Eqn. (15).

$$\lim_{\rho\to\infty}\mathbf{I}_{\phi\phi}^* = \begin{bmatrix} I_{Mv} & 0 \\ 0 & I_{Mf} + I_{\theta\theta}^* \end{bmatrix} \quad (15)$$

In such a design configuration ($\rho\to\infty$), the two input systems are virtually decoupled w.r.t. reflected inertias. This can be recognized from the diagonal structure of the matrix $\lim_{\rho\to\infty}\mathbf{I}_{\phi\phi}^*$.

### A. Physical Meaning of $\mu$ for Designer

In the previous section we have presented an analytical framework that can serve as a design guideline in the consideration of reflected inertias for PFVA driven systems. Having formulated the analytics, an important question we wish to ask is: what is the practical significance of this result?

To answer this question, we will consider the inertia torque demands on the two sub-systems of a PFVA using the inverse dynamics formulation for a 1-DOF single-link equivalent mechanism (as depicted in Fig 3). This is presented in Eqn. (16). As our focus is on dynamic coupling, the non-inertial torque demands (viz. centripetal/Coriolis terms, gravity, and static loads) are not included.

$$\tau_v^I = \left[I_{Mv} + I_{\theta\theta}^* \frac{1}{(\rho+1)^2}\right]\ddot{\phi}_v + \left[I_{\theta\theta}^* \frac{\rho}{(\rho+1)^2}\right]\ddot{\phi}_f$$

$$\tau_f^I = \left[I_{\theta\theta}^* \frac{\rho}{(\rho+1)^2}\right]\ddot{\phi}_v + \left[I_{Mf} + I_{\theta\theta}^* \frac{\rho^2}{(\rho+1)^2}\right]\ddot{\phi}_f \quad (16)$$

Now, we will perform a limit analysis on this set of dynamic equations, based on our findings from Eqn. (15), to obtain Eqn. (17).

$$\lim_{\rho\to\infty}\tau_v^I = [I_{Mv}]\ddot{\phi}_v + [0]\ddot{\phi}_f$$
$$\lim_{\rho\to\infty}\tau_f^I = [0]\ddot{\phi}_v + [I_{Mf} + I_{\theta\theta}^*]\ddot{\phi}_f \quad (17)$$

With regard to reflected inertia, Eqn. (17) suggests that, the force subsystem manages almost all of the output inertia while the velocity subsystem does not see any reflected inertia. This is entirely desirable from a design point of view because a design configuration with a relatively large value of ρ decouples the two input subsystems (FA and VA) w.r.t. inertias. However, there is a practical limitation to achieving this, as described in Section. VII.

## V. NUMERICAL EXAMPLE

The objective of this simulation was to illustrate the analytical formulation presented in this paper, i.e., the effect of relative scale factor (ρ) between the force and velocity inputs of the PFVA on their dynamic coupling (μ). There are two specific issues we intend to illustrate using this simulation; (a) Which input predominantly feels the inertia?, and (b) How does the dynamic coupling term (off-diagonal term) change with ρ? For this simulation, a crank-slider mechanism with a PFVA input was used. The simulation was set up such that the prescribed motion would have appreciable dynamics. A trapezoidal motion plan (constant acceleration – constant velocity – constant deceleration) was imposed on the slider (Fig 5). The initial and final positions of the slider were 0.3263m and 0.5873m, respectively. The maximum acceleration and velocity used for this run were, respectively, 0.3 ms$^{-1}$ and 1 ms$^{-2}$. The relative scaling ratio, ρ, between the force prime-mover and the velocity prime-mover was varied from 5.0 (Relatively Coupled Inputs) to 15.0 (Relatively Uncoupled Inputs).

The numerical result is shown in Fig 5. As shown in this figure, the magnitude of the dynamic coupling term decreases as the relative scale factor increases. Physically, this may be interpreted as follows. If the two sub-systems driving the two inputs of the PFVA have comparable gear ratios, then a predominant part of the inertia torque of each sub-system is used to fight the acceleration of the other sub-system. Hence, as a design guideline, *if the inputs to a PFVA are near-ideal "force" and "velocity" actuators, then this (dynamic) coupling is negligible.*

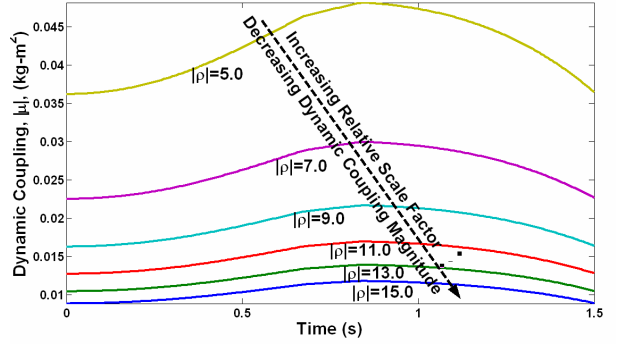

Fig 5. The Effect of Relative scale factor (ρ) on Dynamic Coupling (μ) between the Input Sub-Systems during a Trapezoidal Motion Profile.

## VI. DISCUSSION

In this paper we have presented the concept of a Parallel Force/Velocity Actuator (PFVA) that is based on a Dual Input Single Output (DISO) epicyclic gear train. One of the force paths in this gear train is a high-gear-ratio path and the other one is a low-gear-ratio path. The former input (on the carrier of the gear train), called a "Velocity" input, is not backdriveable and is capable of managing the output velocity without being disturbed by external forces. The latter sub-system (connected to the sun of the planetary gear train), called a "Force" input, is a near-direct drive input that is highly responsive and capable of being backdriven.

An analytical formulation to study the dynamic coupling term (μ) in the apparent inertia matrix at the PFVA input was studied. This was based on an equivalent single-link model for a 1-DOF nonlinear mechanism. Specifically, the variation of μ w.r.t the relative scale factor or ρ (ratio of gear ratios of the force and velocity inputs) was studied. A limit analysis (ρ→∞) was carried out on the apparent inertia matrix to understand the limiting dynamics. It was shown that in the limit (ρ→∞) the two inputs to the PFVA were decoupled w.r.t. inertial torques. In other words:

- *The velocity input does not see a significant apparent inertia.*
- *The force input manages almost all of the inertia of the output.*

The analytical formulation was illustrated by means of a numerical simulation with a crank-slider mechanism incorporating a PFVA input. Results from this simulation corroborated the analytical formulation. It was observed that as the relative scale factor (represented by ρ) was decreased (i.e. the sub-systems tend towards behaving as "equal" systems) the dynamic coupling between the systems increased. Physically, this means that a PFVA design configuration where the two sub-systems have comparable gear ratios has almost no utility since the inertia torque of one of the inputs is predominantly consumed to accelerate the other, thus reducing the actuator's overall efficiency.

## VII. FUTURE WORK

In this section we will present two limitations that are not considered in this paper which are suggested for future work, (1) the physical limitation on the choice of a high value for ρ, and (2) a caveat on the disturbance of the force sub-system (FSS) due to the operation of the velocity sub-system (VSS).

### A. Physical Limitations on the Choice of ρ

The physical limitations on the choice of a high value for the Relative Scale Factor (ρ) arise from the geometry of the planetary gear train. It is necessary to study this issue in greater detail.

### B. Disturbance Analysis of FSS Due to VSS

In our discussion of the reflected inertia issue (Section. IV), we do not consider the disturbance imposed on the force sub-system due to the operation of the velocity sub-system. To address this issue, we suggest that a spring-mass-damper model of the FSS and VSS be considered such that the damping and spring-rate (respectively representing viscous friction and stiffness) are formulated in terms of their (FSS and VSS) gear ratios.

## VIII. ACKNOWLEDGEMENTS

This research was partially funded by the Department of Energy (DOE) grant no. DE-FG52-2004NA25591, under the University Research Program in Robotics (URPR).